\pdfoutput=1

\documentclass[11pt]{article}

\usepackage[final]{acl}

\usepackage{times}
\usepackage{latexsym}

\usepackage[T1]{fontenc}

\usepackage[utf8]{inputenc}

\usepackage{microtype}

\usepackage{inconsolata}

\usepackage{graphicx}
\usepackage{subfigure}
\usepackage{booktabs} 
\usepackage{outlines}
\usepackage{subcaption}
\usepackage{wrapfig}

\definecolor{lightblue}{RGB}{200, 230, 255} 
\definecolor{lightred}{RGB}{255, 210, 210}

\usepackage{adjustbox}
\usepackage{booktabs}

\usepackage{hyperref}

\usepackage{amsmath}
\usepackage{amssymb}
\usepackage{mathtools}
\usepackage{amsthm}
\usepackage{multirow}

\usepackage[capitalize,noabbrev]{cleveref}

\theoremstyle{plain}

\theoremstyle{definition}

\theoremstyle{remark}

\usepackage[textsize=tiny]{todonotes}

\title{Forgotten Polygons: Multimodal Large Language Models are Shape-Blind}

\author{
    William Rudman$^{\star 1}$, Michal Golovanevsky$^{\star 1}$, Amir Bar$^{2}$, Vedant Palit$^{3}$, \\
    {\bf Yann LeCun$^{4}$}, {\bf Carsten Eickhoff$^{5}$}, {\bf Ritambhara Singh$^{1}$} \\
    $^{1}$Brown University, $^{2}$Tel Aviv University, $^{3}$IIT Kharagpur, \\
    $^{4}$New York University, $^{5}$University of Tübingen \\
    \texttt{\{william\_rudman, michal\_golovanevsky\}@brown.edu} \\
}

\begin{document}
\maketitle
\begin{abstract}
\renewcommand{\thefootnote}{\fnsymbol{footnote}}

Despite strong performance on vision-language tasks, Multimodal Large Language Models (MLLMs) struggle with mathematical problem-solving, with both open-source and state-of-the-art models falling short of human performance on visual-math benchmarks. To systematically examine visual-mathematical reasoning in MLLMs, we (1) evaluate their understanding of geometric primitives, (2) test multi-step reasoning, and (3) explore a potential solution to improve visual reasoning capabilities. Our findings reveal fundamental shortcomings in shape recognition, with top models achieving under 50\% accuracy in identifying regular polygons. We analyze these failures through the lens of dual-process theory and show that MLLMs rely on System 1 (intuitive, memorized associations) rather than System 2 (deliberate reasoning). Consequently, MLLMs fail to count the sides of both familiar and novel shapes, suggesting they have neither learned the concept of ``sides'' nor effectively process visual inputs. Finally, we propose Visually Cued Chain-of-Thought (VC-CoT) prompting, which enhances multi-step mathematical reasoning by explicitly referencing visual annotations in diagrams, boosting GPT-4o’s accuracy on an irregular polygon side-counting task from 7\% to 93\%. Our findings suggest that System 2 reasoning in MLLMs remains an open problem, and visually-guided prompting is essential for successfully engaging visual reasoning. Code available at: \href{https://github.com/rsinghlab/Shape-Blind}{
\textit{https://github.com/rsinghlab/Shape-Blind}} \footnote{Equal contribution. Order determined by coin flip.}. 

\end{abstract}

\section{Introduction}
\begin{figure}[htbp]
        \centering
        \includegraphics[width=0.9\columnwidth]{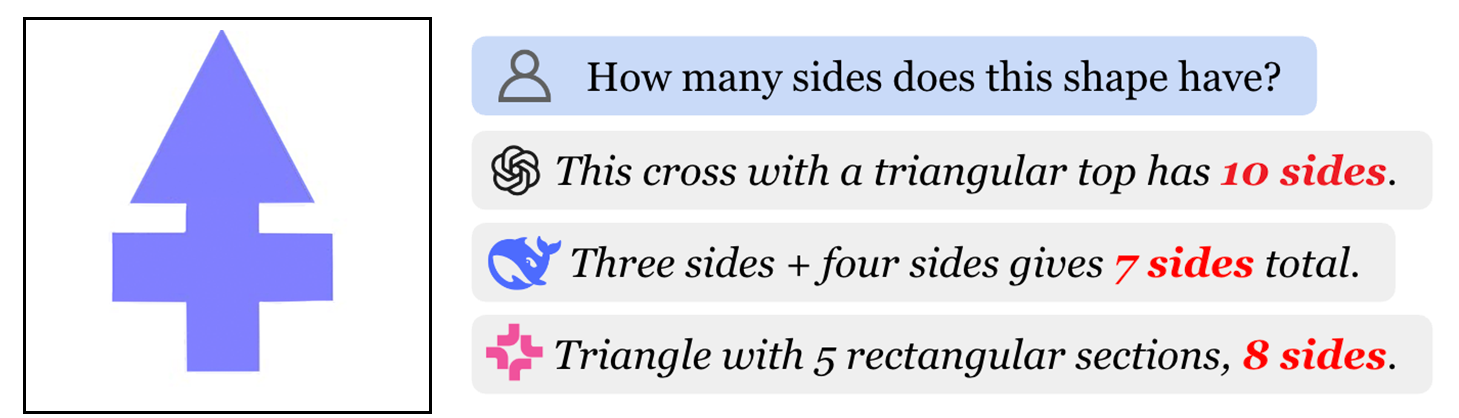} 
        \caption{Multimodal Large Language Models (MLLMs), such as GPT-4o, Janus-Pro, and Molmo fail at counting the number of sides of novel shapes.}
        \label{fig:example_first_page}
    \end{figure}
 
As Multimodal Large Language Models (MLLMs) demonstrate success in vision-language tasks \cite{molmo, janus-pro}, researchers seek to analyze their underlying mechanisms \cite{notice,jiang2024interpreting,luo2024task} and evaluate their ability to perform generalized problem-solving compared to human reasoning \cite{cherian2024evaluatinglargevisionandlanguagemodels}. A key aspect of human reasoning is the interplay between intuitive, reflexive thinking (known as System 1) and deliberate, logical reasoning (known as System 2) \cite{dual-process-1, dual-process-2, fast-slow}. While humans engage in both types of reasoning, a key question arises: \emph{Can MLLMs move beyond memorized responses of System 1 to engage in the analytical reasoning of System 2?}

Mathematics provides a powerful framework for this investigation, as fundamental reasoning skills are crucial for complex problem-solving. The combination of logic, abstract concepts, and specialized symbolic language makes mathematical reasoning problems a strong test of the extent to which models engage in System 2 reasoning rather than relying solely on memorized concepts \cite{satpute2024taxonomy}. Studying System 2 reasoning abilities is especially relevant given recent progress in natural language processing, where Large Language Models (LLMs) like GPT-4 \cite{achiam2023gpt}, Qwen \cite{bai2023qwen}, and LLaMA \cite{touvron2023llama} have demonstrated notable success on mathematical benchmarks like GSM8K \cite{cobbe2021training} and MATH \cite{hendrycks2021measuring}. Despite these advancements, MLLMs remain limited in visual-mathematical reasoning. While closed-source models like GPT-4o demonstrate potential, their performance on benchmarks such as MathVerse \cite{zhang2024mathversedoesmultimodalllm} remains far below human levels, with open-source models typically performing even worse~\cite{zhang2024mathversedoesmultimodalllm}. Even MLLMs specially trained for visual, mathematical reasoning (G-LLaVA \cite{g-llava}, Math-PUMA \cite{mathpuma}, Math-LLaVa \cite{math-llava} cannot close the gap between open-source and closed-source models.
Many benchmarks feature complex geometric diagrams that require multi-step reasoning, with human performance reaching no more than 70\% accuracy \cite{mathvision, zhang2024mathversedoesmultimodalllm}. \citet{cherian2024evaluatinglargevisionandlanguagemodels} show that MLLMs tend to perform better on more complex mathematical tasks but struggle with simple tasks designed for young children. This counterintuitive result raises concerns about whether MLLM performance on visual-mathematical benchmarks reflects genuine mathematical reasoning (System 2) or merely the retrieval of familiar concepts from training data (System 1). 

We design a two-part task to evaluate MLLMs' reasoning abilities: (1) shape recognition, which relies on visual recall (System 1), and (2) side counting, which engages visual reasoning (System 2). Starting with common geometric shapes and properties allows us to pinpoint specific failures in how MLLMs process mathematical diagrams. Our results show that while the underlying LLMs demonstrate perfect accuracy on non-visual questions about polygon names and side counts, their multimodal counterparts fail when models have to rely on images to answer questions correctly. Unlike humans, who can count a shape's sides and infer its identity (e.g., ``1, 2, \dots, 7'' $\rightarrow$ ``heptagon''), our results suggest that MLLMs do not attempt such reasoning. Instead, they rely on System 1 heuristics like shape memorization and spurious correlations between visual features and labels, ignoring vital information in the image. 

To understand why MLLMs struggle with polygon identification from visual inputs, we analyze vision encoder embeddings. Our results show that vision encoders are ``shape-blind'':common shapes form distinct clusters, while less frequent ones like pentagons, heptagons, and octagons overlap (Section~\ref{exp1_vision_only}). Even models specifically trained for geometric understanding, such as G-LLaVA, Math-LLaVA, and Math-PUMA, exhibit shape-blindness where distinct polygons are embedded in the same region in vector space. 
We further demonstrate vision encoders' shape-blindness with a two-shape multi-step task: identifying shapes, counting the sides, and computing the total sum of sides, adding complexity by incorporating arithmetic into the reasoning process. On average, models correctly identify both shapes in an image only 27.58\% of the time, as shape recognition depends entirely on visual input. However, MLLMs compute the sum of the identified shapes' sides correctly in 70.75\% of cases, indicating that despite errors in recognition, the summation operation remains highly accurate. While open-source models struggle with less common polygons, GPT-4o identifies shapes with at least 49\% accuracy. To test whether models use System 2 reasoning, we introduce a dataset of abstract shapes and irregular polygons unlikely to have appeared in training (see Figure \ref{fig:example_first_page}). No model accurately counts the sides of these novel shapes, reinforcing their reliance on memorization rather than genuine geometric understanding.

As MLLMs continue to advance, our findings urge the community to re-examine the complexity of visual-mathematical benchmarks. If state-of-the-art open-source models fail to recognize simple shapes, should we be evaluating them on complex geometric tasks? Our results suggest that current models rely heavily on factual recall rather than true geometric reasoning, limiting their ability to generalize beyond familiar shapes. To explore ways of bridging this gap, we take a step forward with \textit{Visually-Cued} Chain-of-Thought (VC-CoT) prompting, leveraging the fact that many geometric datasets already contain annotated shapes (e.g., labeled triangle vertices A, B, and C). By explicitly guiding the model to reference image annotations (letters or numbers) and reason about their relationships, VC-CoT significantly improves GPT-4o’s ability to count the sides of novel shapes, boosting accuracy from 7\% to 93\%. Across Molmo, Janus-Pro, GPT-4-Turbo, and GPT-4o, our VC-CoT prompts achieve an average accuracy improvement of approximately 5\% compared to standard CoT prompts on MathVerse. These findings highlight the importance of structured prompting in strengthening the connection between visual perception and mathematical reasoning in MLLMs.

\section{Related Works}
Despite the success of MLLMs, emerging research highlights their struggles with reasoning.~\citet{alhamoud2025vision} find that vision-language models struggle to negate in retrieval and multiple-choice tasks. Beyond general reasoning, open-source, closed-source, and math-specific MLLMs perform poorly on visual math benchmarks \cite{mathvision, zhang2024mathversedoesmultimodalllm}. 

A key limitation in MLLMs is the vision encoder. \citet{clip-blind} show CLIP-based vision encoders fail to capture fine-grained details \textemdash a detrimental trait for tasks requiring precise visual reasoning. Similarly, in the visual-mathematics domain, \citet{g-llava, math-llava, mavis, mathpuma} demonstrate that vision encoders produce inadequate representations of mathematical diagrams and vision-text misalignments  hinder multimodal reasoning.

One way to address these challenges is fine-tuning. Recent works adopt a multi-step approach: (1) fine-tuning vision encoders to enhance visual representation, (2) training modality projectors for better alignment, and (3) instruction-tuning with CoT datasets \cite{g-llava, math-llava, mavis, mathpuma}. While this may increase performance on math benchmarks, studies show that fine-tuning foundation models often reduces generalization ability and does not address overarching reasoning capability \cite{generalization_fine_tuning, sft_generalize}.

In the language domain, Chain-of-Thought (CoT) prompting has proven highly effective in encouraging System 2 reasoning in LLMs \cite{cot}. \citet{xiang2025towards} further reinforces this, demonstrating that structuring reasoning through CoT can significantly improve logical inference and problem-solving. However, applying CoT to MLLMs has been far less successful. \citet{vlm_cot} show that open-source MLLMs struggle with CoT, largely due to limitations in existing visual-instruction tuning datasets, which prioritize short, simplistic responses over structured reasoning. Thus, recent works explicitly fine-tune MLLMs for CoT reasoning, particularly in object counting and mathematical reasoning tasks \cite{vlm_cot, mavis, mathpuma, molmo}. However, even these fine-tuned models continue to struggle on complex visual-mathematical benchmarks such as MathVista \cite{math-vista}, and MathVerse \cite{zhang2024mathversedoesmultimodalllm}. These challenges highlight that despite recent efforts, System 2 reasoning in MLLMs remains an open problem, with current approaches failing to achieve generalizable reasoning abilities.

\section{Experiments}
\textbf{Models Evaluated:} In our experiments, we evaluate 13 diverse MLLMs. We consider (1) general open-source models, including LLaVA-1.5-7B \cite{llava-1.5}, LLaVA-Next-7B \cite{llava-next}, LLaVA-OneVision-7B \cite{llava-onevision}, Qwen2-VL-7B \cite{qwen2-vl}, InternVL-8B \cite{qwen2-vl}, Molmo-7B \cite{molmo}, and DeepSeek’s Janus Pro-7B \cite{janus-pro}; (2) math-specialized open-source models, including Math-LLaVA-13B \cite{math-llava}, G-LLaVA-7B \cite{g-llava}, and Math-PUMA-7B \cite{mathpuma}; and (3) closed-source models, including GPT-4-Turbo and GPT-4o. See Table~\ref{tab:model_details} in Appendix \ref{app:models} for full evaluation details.

\subsection{Probing Geometric Knowledge}
\label{exp1_mllms} 
 \begin{table*}[h]
     \centering
    \scriptsize
    \setlength{\tabcolsep}{9pt}
    \begin{tabular}{lcccccc}
        \toprule
        \multicolumn{7}{c}{\textbf{What shape is in the image? / How many sides does the shape in the image have?}} \\
        \midrule
        Model & Triangle & Square & Pentagon & Hexagon & Heptagon & Octagon \\
        \midrule
        LLaVA-1.5 & 96\% / 93\% & 99\% / 100\% & \textcolor{red}{0\%} / \textcolor{red}{0\%} & 5\% / 90\% & \textcolor{red}{0\%} / \textcolor{red}{0\%} & \textcolor{red}{0\%} / 3\% \\
        LLaVA-Next & \textbf{100\%} / 99\% & \textbf{100\%} / 98\% & \textcolor{red}{0\%} / \textcolor{red}{0\%} & 94\% / \textcolor{red}{0\%} & \textcolor{red}{0\%} / \textcolor{red}{0\%} & 26\% / \textbf{99\%} \\
        LLaVA-OneVision & \textbf{100\%} / \textbf{100\%} & \textbf{100\%} / \textbf{100\%} & 54\% / 72\% &  \textbf{100\%} / \textbf{100\%} & \textcolor{red}{0\%} / \textcolor{red}{0\%} & \textbf{100\%} / \textbf{99\%} \\
        Qwen2-VL & \textbf{100\%} / \textbf{100\%} & \textbf{100\%} / \textbf{100\%} & 97\% / 99\% & 92\% / 88\% & 14\% / 3\% & 99\% / 97\% \\
        LLaMA-3.2 & \textbf{100\%} / 98\% & 99\% / 78\% & 85\% / 70\% & 94\% / \textbf{96\%} & \textcolor{red}{0\%} / 9\% & 42\% / 20\% \\
        InternVL & \textbf{100\%} / \textbf{100\%} & 98\% / \textbf{100\%} & \textcolor{red}{0\%} / 80\% & 89\% / 77\% & \textcolor{red}{0\%} / \textcolor{red}{0\%} & 5\% / 3\% \\
        Molmo & \textbf{100\%} / \textbf{100\%} & 99\% / \textbf{100\%} & 47\% / 7\% & 93\% / 94\% & 55\% / 51\% & 87\% / 94\% \\
        Janus-Pro & \textbf{100\%} / \textbf{100\%} & \textbf{100\%} / \textbf{100\%} & \textbf{100\%} / \textbf{100\%} & \textbf{100\%} / \textbf{100\%} & \textcolor{red}{0.6\%} / 4\% & 65\% / 77\% \\
        \midrule
        Math-LLaVA & \textbf{100\%} / 86\% & \textbf{100\%} / \textbf{100\%} & 42\% / 79\% & 78\% / 61\% & \textcolor{red}{0\%} / \textcolor{red}{0\%} & 87\% / 90\% \\
        G-LLaVA & 93\% / 12\% & 50\% / 92\% & \textcolor{red}{0\%} / 8\% & 55\% / 92\% & \textcolor{red}{0\%} / \textcolor{red}{0\%} & \textcolor{red}{0\%} / 9\% \\
        Math-PUMA & \textbf{100}\% / \textbf{100\%} & \textbf{100\%} / \textbf{100\%} & 26\% / 18\% & \textbf{100\%} / \textbf{100\%} & \textcolor{red}{0\%} / \textcolor{red}{0\%} & \textcolor{red}{0\%} / \textcolor{red}{0\%} \\
        \midrule
        GPT-4-Turbo & \textbf{100\%} / 98\% & 98\% / 99\% & 87\% / 85\% & \textbf{100\%} / 95\% & \textcolor{red}{0.1\%} / 10\% & \textbf{69\%} / 78\% \\
        GPT-4o & 99\% / \textbf{100\%} & 97\% / \textbf{100\%} & \textbf{100\%} / \textbf{100\%} & 68\% / 78\% & \textbf{92\%} / 58\% & 49\% / 71\% \\
        \bottomrule
    \end{tabular}
    \caption{\textbf{Model accuracy for shape identification (first value) and side counting (second value)}. Notably, models struggle to count sides accurately. GPT-4o identifies heptagons with 92\% accuracy but counts sides correctly only 58\% of the time, revealing a gap between recognition and reasoning.}
    \label{tab:combined_shape_sides_accuracy}
\end{table*}
Geometric reasoning provides a natural testbed for distinguishing between System 1 heuristics, such as memorization of common shape patterns, and System 2 reasoning, which requires deliberate reasoning, such as counting sides of a shape. To examine this distinction, we create a dataset comprising six common regular polygons: triangle, square, pentagon, hexagon, heptagon, and octagon (see examples in Figure \ref{fig:regular_polygons} in Appendix \ref{app:one_shape}). For each shape, we generate images with different colors, rotations, and sizes, creating a dataset of 2000 images. All images are 400×400 pixels, and we ensure that even the smallest shapes occupy at least 15\% of the image (more details in Appendix Section \ref{sec:image_gen}). Our analysis follows a three-stage approach to disentangle the contributions of vision, language, and their integration in MLLMs. First, we assess the MLLM as a whole, evaluating its ability to connect visual information to geometric properties (Section \ref{exp1_mllms}). Next, we prompt the underlying LLMs in a text-only setting to determine their knowledge independent of vision (Section \ref{exp1_llms_only}). Finally, we analyze the vision encoder by examining the vision embeddings, shedding light on what the model ``sees'' without language guidance (Section \ref{exp1_vision_only}).

For each image, we use two prompts to test for geometric knowledge: ``What shape is in the image?'' and ``How many sides does the shape in the image have?''. Table~\ref{tab:combined_shape_sides_accuracy} shows that MLLMs excel at identifying common shapes like triangles and squares, with many models achieving perfect accuracy. However, performance drops sharply for less familiar polygons such as pentagons, heptagons, and octagons. Most models fail on heptagons, with only GPT-4o and Molmo exceeding 1\% accuracy. Notably, Molmo is trained with a point-then-count approach, identifying objects before quantifying them. Despite this targeted training, Molmo correctly counts pentagon sides only 7\% of the time and heptagons 51\% of the time (more examples of Molmo behavior in Appendix~\ref{app:molmo_experiment}). Moreover, GPT-4o correctly identifies heptagons 92\% of the time but counts their sides accurately only 58\% of the time, revealing its tendency to recognize shapes without using visual information to reason about their properties. These findings prompt a key question: is the breakdown in shape recognition and side-counting due to flawed visual processing or a failure to apply reasoning? We explore this by analyzing the vision encoder and LLM backbone separately.

\subsubsection{Which visual features affect accuracy?}\label{exp1_correlation_analysis}
To better understand the causes of model errors, we analyzed the relationship between accuracy and five visual attributes: background color, shape color, size, rotation, and contrast ratio. The contrast ratio is defined as the difference between the background and shape colors. Table~\ref{tab:correlation_table} shows the correlation coefficients between each factor and model accuracy for thirteen different models. A positive correlation means that accuracy improves as the factor increases, while a negative correlation suggests the factor has a harmful effect on accuracy.

\begin{table}[h]
    \centering
    \tiny
    \setlength{\tabcolsep}{4pt}
    \begin{tabular}{lccccc}
        \toprule
        \textbf{Model} & \textbf{\shortstack{Background\\Color}} & \textbf{\shortstack{Shape\\Color}} & \textbf{Size} & \textbf{Rotation} & \textbf{\shortstack{Contrast\\Ratio}} \\
        \midrule
        LLaVA 1.5 & -0.005 & \textbf{0.066} & 0.126 & 0.089 & 0.047 \\
        LLaVA 1.6 & -0.197 & -0.015 & 0.039 & 0.010 & -0.110 \\
        Qwen-VL & 0.127 & 0.009 & 0.096 & -0.226 & 0.062 \\
        InternVL & -0.042 & 0.061 & 0.186 & 0.011 & -0.065 \\
        LLaVA-OneVision & 0.012 & 0.033 & 0.026 & -0.001 & -0.020 \\
        LLaMA 3.2 & 0.021 & 0.026 & 0.049 & 0.010 & 0.014 \\
        Molmo & 0.003 & 0.020 & 0.039 & -0.270 & -0.005 \\
        Janus & -0.184 & 0.008 & \textbf{0.223} & -0.144 & \textbf{-0.146} \\
        GPT-4-Turbo & -0.056 & 0.051 & 0.035 & -0.092 & -0.023 \\
        GPT-4o & 0.031 & 0.039 & -0.054 & -0.144 & 0.068 \\
        Math-LLaVA & 0.122 & -0.001 & -0.006 & \textbf{-0.362} & 0.066 \\
        G-LLaVA & \textbf{0.209} & 0.060 & 0.023 & -0.040 & 0.137 \\
        Math-PUMA & 0.018 & -0.017 & 0.172 & -0.079 & -0.020 \\
        \midrule
        \textbf{Abs Val Avg} & \textbf{0.079} & \textbf{0.031} & \textbf{0.083} & \textbf{0.098} & \textbf{0.060} \\
        \bottomrule
    \end{tabular}
    \caption{Correlation between visual features and model accuracy. Values represent how strongly each factor relates to correct predictions. Rotation and size have the highest average influence.}
    \label{tab:correlation_table}
\end{table}

The results show that no single visual factor explains performance across all models. For example, rotation has a strong negative effect in Qwen2-VL and Math-LLaVA but has almost no effect in LLaVA-OneVision. On average, rotation shows the strongest overall influence, followed by size.

Contrast ratio, a factor that is often important in traditional vision tasks, shows mixed results. GPT-4o and G-LLaVA perform slightly better with higher contrast, while LLaVA 1.6 and Janus perform worse. These inconsistencies suggest that visual sensitivity varies significantly depending on model architecture and training.

Taken together, these findings suggest that misclassifications are not caused by a single weak point in the visual input. Instead, each model exhibits its own unique set of sensitivities. These insights complement our broader evaluation of shape recognition and motivate a closer look at the components of the system. In the next sections, we analyze the language model and the vision encoder independently to understand their contributions to these errors.

\subsubsection{What does the text-decoder know?}\label{exp1_llms_only} \begin{table}[h]
    \centering
    \scriptsize
    \resizebox{\columnwidth}{!}{
    \begin{tabular}{lc}
        \toprule
        \multicolumn{2}{c}\textbf{Name the \textit{ \textless n \textgreater}-sided polygon / Num of sides in \textless \textit{shape} \textgreater} \\
        \midrule
        LM Only & Accuracy (\%) \\
        \midrule
        LLaVA-1.5 & 67\% / 100\% \\
        LLaVA-Next & 100\% / 100\% \\
        LLaVA-OneVision & 100\% / 100\% \\
        Qwen2-VL & 100\% / 100\% \\
        LLaMA-3.2 & 100\% / 100\% \\
        InternVL & 100\% / 100\% \\
        Molmo & 100\% / 100\% \\
        Janus-Pro & 100\% / 100\% \\
        \midrule
        Math-LLaVA & 100\% / 100\% \\
        G-LLaVA & 50\% / 100\% \\
        Math-PUMA & 100\% / 100\% \\
        \midrule
        GPT-4o & 100\% / 100\% \\
        GPT-4-Turbo & 100\% / 100\% \\
        \bottomrule
    \end{tabular}
    }
    \caption{Accuracy of the LLM-backbone on the polygon naming and side counting tasks: 
    ``What is the name of a \textit{\textless n\textgreater}-sided regular polygon? / 
    How many sides does \textit{\textless shape\textgreater} have?'' for $n \in \{3,4,...,8 \}$ and shape in $\{$triangle, square,..., octagon$\}$.} 
    \label{tab:LM_only}
\end{table}
To assess LLMs’ knowledge of the geometric properties MLLMs struggle with, we construct the following prompts: (1) ``What is the name of a \textit{\textless n\textgreater}-sided polygon?'' where \textit{\textless n\textgreater} ranges from 3 to 8, and (2) ``How many sides does a \textit{\textless shape\textgreater} have?'' where \textit{\textless shape\textgreater} is a regular polygon from triangle to octagon. The results, shown in Table~\ref{tab:LM_only}, reveal that the underlying language models know geometric properties. All models get 100\% accuracy on ``How many sides does a \textit{\textless shape\textgreater} have?'', and 11 out of 13 models get 100\% accuracy on ``What is the name of a \textit{\textless n\textgreater}-sided polygon?''. The success of LLM backbones in these tasks aligns with prior findings that state-of-the-art models perform well on lower-level math problems \cite{xiang2025towards}. Note that LLaVA 1.5 and G-LLaVA's LMs struggle with naming certain shapes, suggesting some failures stem from gaps in pretraining data rather than vision-related issues. In Appendix~\ref{app:ngrams_analysis}, we analyze Google N-Grams data and confirm that the shapes these models struggle with are also the least frequently mentioned in text. 

\subsubsection{What does the vision-encoder see?}\label{exp1_vision_only}

\begin{figure}[htbp]
        \centering
        \includegraphics[width=0.9\columnwidth]{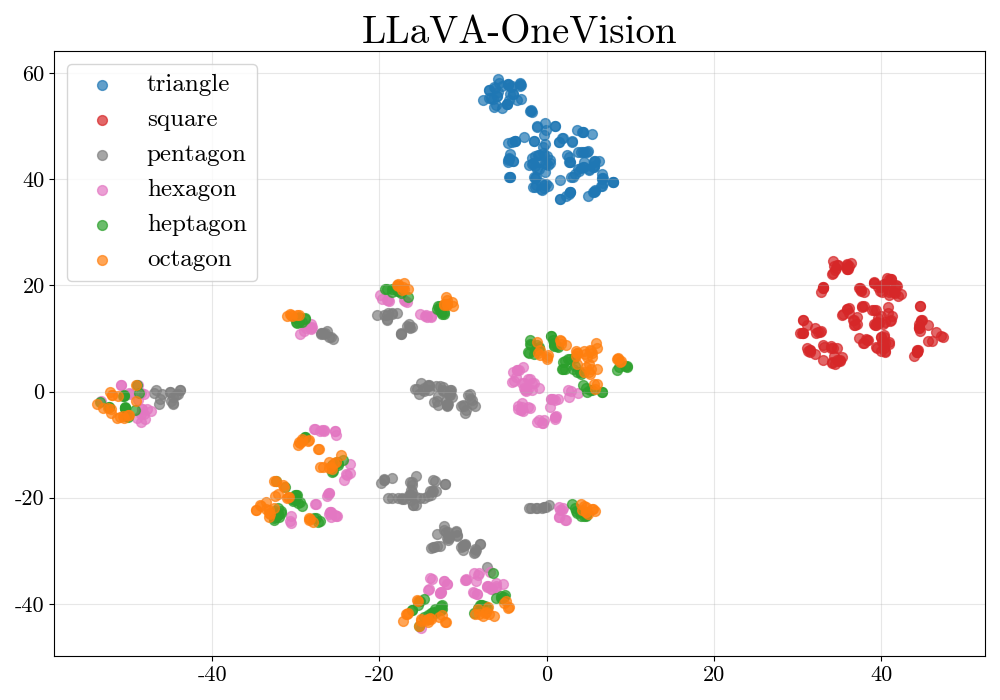} 
        \caption{\textbf{T-SNE plots of vision encoder embeddings from LLaVA-OneVision}. Only triangles and squares form distinct clusters. Appendix~\ref{app:vision_encoder_experiments} shows all models learn a similar embedding.}
        \label{fig:cluster}
    \end{figure}
    
\begin{figure*}[h!]
        \centering\includegraphics[width=\textwidth]{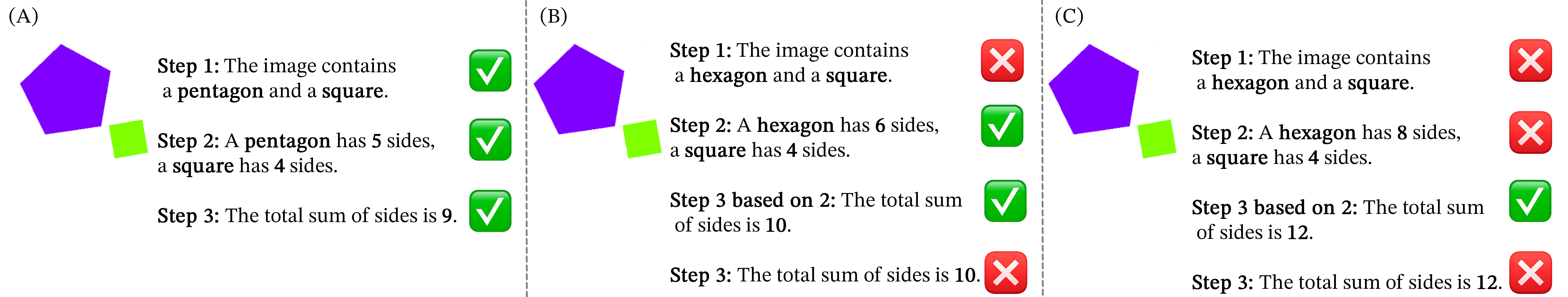} 
        \caption{\textbf{Illustration of failure modes in the two-shape reasoning task}.  
        \textbf{A}: Successful completion of all steps.  
        \textbf{B}: The most common failure mode, where misidentification in Step 1 leads to an incorrect sum in Step 3.  
        \textbf{C}: An error in mapping shapes to their number of sides (Step 2), affecting the final sum.} \label{fig:step1}
    \end{figure*}
To evaluate how vision encoders process shapes, we use t-SNE to visualize their embeddings. Figure~\ref{fig:cluster} shows results for LLaVA-OneVision, which exhibits the same embedding pattern as all other models (Appendix~\ref{app:vision_encoder_experiments}). Common shapes like triangles (blue) and squares (red) form well-defined clusters, reflecting the models' strong performance on these shapes (Table~\ref{tab:combined_shape_sides_accuracy}). In contrast, vision encoders appear ``shape-blind'' to less common shapes, often embedding them into the same cluster. In particular, hexagons (pink), heptagons (green), and octagons (orange) exhibit dispersed and overlapping embeddings indicating poor differentiation. Appendix~\ref{app:vision_encoder_experiments} provides nearest neighbor analysis for all models, confirming that triangles and squares form distinct clusters in vision-encoder representations. In LLaVA-OneVision, 99.1\% of a triangle's 20 nearest neighbors are also triangles, while only 41.1\% of a heptagon's neighbors are heptagons, with most being hexagons or octagons. This trend is consistent across all open-source models. 

Although LLMs excel at answering basic shape property questions, vision encoders in MLLMs struggle to differentiate shapes, causing the entire MLLM system to misidentify shapes. The contrast between Table~\ref{tab:combined_shape_sides_accuracy} and Table~\ref{tab:LM_only} suggests vision encoder limitations as a key failure, aligning with prior findings on their lack of fine-grained detail \cite{nayak2022learning}. The interplay between weak vision encoders and strong LLM backbones is evident in GPT-4o and GPT-4 Turbo. GPT-4o outperforms GPT-4-Turbo in shape identification, raising heptagon identification from 0\% to 92\%, yet struggles with side counting (58\% accuracy). While a human would engage in System 2 reasoning and count the sides (e.g., ``1, 2, 7'' $\rightarrow$ heptagon) when faced with an unfamiliar shape, models instead default to predicting 6 or 8 sides, suggesting reliance on memorized patterns (System 1) rather than genuine visual reasoning. These findings highlight a fundamental issue: recognizing shapes in pretraining data does not generalize to an understanding of geometric properties such as having ``sides''. This disconnect prompts an examination of how the lack of visual understanding of ``sides'' impacts performance on more complex, multi-step reasoning tasks, which more closely resemble real-world math datasets.

\subsection{Evaluating Multi-step Math Reasoning}

\begin{table}[h]
\scriptsize
    \resizebox{\columnwidth}{!}{
    \centering
    \begin{tabular}{lcccc}
        \toprule
        \multicolumn{5}{c}{\textbf{1. Identify the shapes, 2. Map shape to side, 3. Sum all sides.}} \\
        \midrule
        VLM Accuracy & Step 1 & Step 2  & Step 3 from 2 & Step 3 \\
        \midrule
        LLaVA-1.5 & 8.92\% & 96.94\% & 87.50\% & 11.61\% \\
        LLaVA-Next & 14.09\% & 99.80\% & 97.76\% & 26.33\% \\
        LLaVA-OneVision & 53.38\% & 99.98\% & \textbf{99.96\%} & 43.61\% \\
        Qwen2-VL & 71.71\% & \textbf{100.00\%} & 98.70\% & 55.15\% \\
        LLaMA-3.2 & 2.67\% & 99.33\% & 38.25\% & 31.02\% \\
        InternVL & 41.63\% & 99.88\% & 98.13\% & 43.54\% \\
        Molmo & 27.92\% & 97.35\% & 99.92\% & 27.27\% \\
        Janus-Pro & 50.77\% & 95.05\% & 63.26\% & 52.20\% \\
        \midrule
        Math-LLaVA & 0.00\% & 0.06\% & 0.00\% & 14.18\% \\
        G-LLaVA & 0.00\% & 0.00\% & 0.00\% & 3.81\% \\
        Math-PUMA & 32.08\% & \textbf{100.00}\% & 78.29\% & 33.32\% \\
        \midrule
        GPT-4-Turbo & 52.95\% & 99.96\% & 81.73\% & 50.59\% \\
        GPT-4o & \textbf{73.40\%} & \textbf{99.98\%} & 97.74\% & \textbf{56.23\%}\\
        \bottomrule
    \end{tabular}
}
    \caption{\textbf{Step-wise accuracies}. Although models incorrectly identify the shapes, they still perform the correct mapping of shapes to their respective sides and then correctly sum. See Figure~\ref{fig:step1}.}
    \label{tab:stepwise_accuracy}
\end{table}

We use a three-step mathematical reasoning pipeline to pinpoint bottlenecks in MLLM geometric reasoning: Step 1: identifying shapes in an image; Step 2: retrieving their side counts; and Step 3: summing the sides of two shapes. Combining multi-shape recognition with arithmetic operations helps separate vision challenges (e.g., shape recognition) from reasoning limitations (e.g., arithmetic errors or mapping mistakes). Following the single-shape generation procedure (Appendix~\ref{app:two_shape}), we extend the dataset to include two-shape images with variations in color, background, size, and rotation, ensuring no overlap or collision.
\begin{table*}[h]
    \resizebox{\textwidth}{!}{ 
        \begin{tabular}{lcccccccccccccc}
            \toprule
    Model & Tri-Sq & Tri-Pen & Tri-Hex & Tri-Hep & Tri-Oct & Sq-Pen & Sq-Hep & Irr-Poly & Arrow & Star & Plus & Arrow-Plus \\
    \midrule
    LLaVA-1.5 & \textcolor{red}{0\%} & \textcolor{red}{0\%} & 10\% & 30\% & \textcolor{red}{0\%} & \textcolor{red}{0\%} & \textcolor{red}{0\%} & 10\% & \textcolor{red}{0\%} & \textcolor{red}{0\%} & \textcolor{red}{0\%} & \textcolor{red}{0\%} \\
    LLaVA-Next & \textcolor{red}{0\%} & \textcolor{red}{0\%} & 10\% & \textcolor{red}{0\%} & \textcolor{red}{0\%} & \textcolor{red}{0\%} & \textcolor{red}{0\%} & 10\% & \textcolor{red}{0\%} & \textcolor{red}{0\%} & \textcolor{red}{0\%} & \textcolor{red}{0\%} \\
    LLaVA-OneVision & \textcolor{red}{0\%} & \textcolor{red}{0\%} & \textbf{100\%} & \textcolor{red}{0\%} & \textcolor{red}{0\%} & \textcolor{red}{0\%} & \textcolor{red}{0\%} & 36\% & \textcolor{red}{0\%} & \textcolor{red}{0\%} & \textcolor{red}{0\%} & \textcolor{red}{0\%} \\
    Qwen2-VL & \textcolor{red}{0\%} & 20\% & 40\% & 40\% & \textcolor{red}{0\%} & \textcolor{red}{0\%} & \textcolor{red}{0\%} & 19\% & \textcolor{red}{0\%} & \textcolor{red}{0\%} & \textcolor{red}{0\%} & \textcolor{red}{0\%} \\
    InternVL & 10\% & \textcolor{red}{0\%} & \textcolor{red}{0\%} & \textcolor{red}{0\%} & \textcolor{red}{0\%} & \textcolor{red}{0\%} & \textcolor{red}{0\%} & 18\% & \textcolor{red}{0\%} & \textcolor{red}{0\%} & \textcolor{red}{0\%} & \textcolor{red}{0\%} \\
    LLaMA-3.2 & \textbf{90\%} & \textcolor{red}{0\%} & 40\% & \textcolor{red}{0\%} & \textcolor{red}{0\%} & \textcolor{red}{0\%} & \textcolor{red}{0\%} & 36\% & \textcolor{red}{0\%} & \textcolor{red}{0\%} & \textcolor{red}{0\%} & \textcolor{red}{0\%} \\
    Molmo & 40\% & \textbf{40\%} & 70\% & \textbf{50\%} & \textcolor{red}{0\%} & \textbf{10\%} & \textcolor{red}{0\%} & 60\% & \textbf{42\%} & \textcolor{red}{0\%} & 3\% & \textcolor{red}{0\%} \\
    Janus-Pro & 70\% & \textcolor{red}{0\%} & 70\% & \textcolor{red}{0\%} & \textcolor{red}{0\%} & \textcolor{red}{0\%} & \textcolor{red}{0\%} & 35\% & \textcolor{red}{0\%} & \textcolor{red}{0\%} & \textcolor{red}{0\%} & \textcolor{red}{0\%} \\
    \midrule
    Math-LLaVA & \textcolor{red}{0\%} & \textcolor{red}{0\%} & \textcolor{red}{0\%} & \textcolor{red}{0\%} & \textcolor{red}{0\%} & \textcolor{red}{0\%} & \textcolor{red}{0\%} & \textbf{42\%} & \textcolor{red}{0\%} & \textcolor{red}{0\%} & \textcolor{red}{0\%} & \textcolor{red}{0\%} \\
    G-LLaVA & \textcolor{red}{0\%} & \textcolor{red}{0\%} & 40\% & 44\% & 10\% & \textcolor{red}{0\%} & \textcolor{red}{0\%} & \textbf{100\%} & 23\% & \textcolor{red}{0\%} & \textcolor{red}{0\%} & \textcolor{red}{0\%} \\
    Math-PUMA & \textcolor{red}{0\%} & \textcolor{red}{0\%} & \textcolor{red}{0\%} & \textcolor{red}{0\%} & \textcolor{red}{0\%} & \textcolor{red}{0\%} & \textcolor{red}{0\%} & \textcolor{red}{0\%} & \textcolor{red}{0\%} & \textcolor{red}{0\%} & \textcolor{red}{0\%} & \textcolor{red}{0\%} \\
    \midrule
    GPT-4-Turbo & 50\% & \textcolor{red}{0\%} & 30\% & \textcolor{red}{0\%} & 10\% & \textcolor{red}{0\%} & \textbf{10\%} & 38\% & \textbf{89\%} & 34\% & \textbf{92\%} & \textcolor{red}{0\%} \\
    GPT-4o & 60\% & \textcolor{red}{0\%} & 60\% & \textcolor{red}{0\%} & \textbf{60\%} & \textcolor{red}{0\%} & \textbf{10\%} &  \textbf{42\%} & \textbf{89\%} & \textbf{84\%} & 88\% & \textcolor{red}{0\%} \\
    \bottomrule
        \end{tabular}
        }
\captionof{table}{\textbf{Performance on merged shapes, irregular polygons, and abstract shapes.} (see examples in Figure~\ref{fig:shapes}). GPT-4o and GPT-4-Turbo excel on common shapes ('arrow,' 'star,' 'plus') but fail (0\%) on unseen shapes, suggesting reliance on recognition over side counting.}
        \label{tab:composite_shapes}
\end{table*}
Table~\ref{tab:stepwise_accuracy} reveals that most models struggle in the two-shape setting, with the shape identification step (Step 1) emerging as the primary bottleneck. While models like GPT-4o and GPT-4 Turbo achieve near-perfect accuracy in recalling shape properties (Step 2), their failure to correctly recognize shapes in Step 1 limits overall performance. For instance, GPT-4o identifies shapes with 73.40\% accuracy but achieves only 56.23\% accuracy in Step 3. Similarly, LLaVA-Next, InternVL, LLaVA-OneVision, and Molmo perform well in mapping and arithmetic reasoning but frequently misidentify shapes, leading to errors seen in Figure~\ref{fig:step1}, panel B. In addition to the ground-truth sum of sides, we evaluate accuracy for Step 3 based on the output of Step 2, allowing us to isolate errors in shape identification from arithmetic errors (see Figure~\ref{fig:step1}). Consider an image containing a pentagon and a square. If the model misidentifies the pentagon as a hexagon, Step 3 accuracy would still be correct if the sum aligns with the model's incorrect mapping (e.g., 6 + 4 = 10 instead of the correct 5 + 4 = 9). Table~\ref{tab:stepwise_accuracy} demonstrates that models execute sums correctly based on the side mapping from Step 2. Namely, the low performance in Step 3 accuracies is primarily due to an inability to retrieve the correct shapes in the image, not an inability to execute sums. 

Examining reasoning through distinct steps highlights a fundamental limitation in multimodal integration. MLLMs succeed in basic arithmetic but fail when vision and language must work together. Their tendency to rigidly map less common shapes to familiar ones (e.g., misclassifying heptagons as hexagons) highlights the lack of reliable visual grounding. Even with strong LLM backbones, this limitation persists, reflecting broader challenges in vision-language mathematical benchmarks \cite{zhang2024mathversedoesmultimodalllm}. To investigate this further, we evaluate performance on entirely unseen shapes devoid of familiar visual priors and biases.

\subsection{Counting Sides of Abstract Shapes}
We introduce a dataset of novel shapes to test the hypothesis that MLLMs rely on System 1 reasoning—memorizing shape-side relationships rather than actively counting and using visual information. These shapes are unlikely to appear in training, allowing us to evaluate whether models can generalize beyond familiar regular polygons. We consider three distinct categories of abstract shapes:

\begin{figure}
    \centering
    \includegraphics[width=0.7\columnwidth]{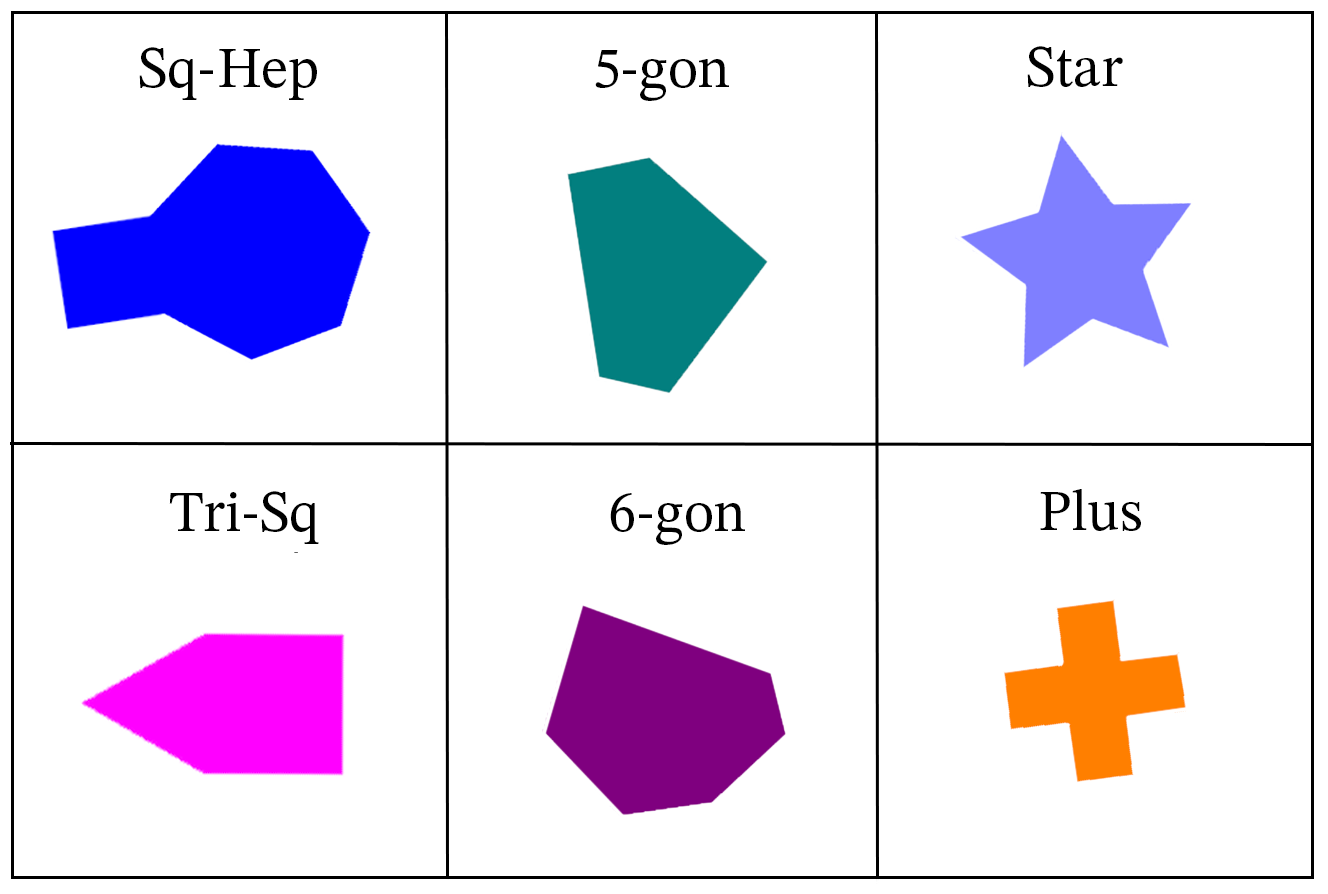}
    \caption{Examples of abstract shapes. For the full set of shapes, see Figure~\ref{app:abstract_shapes} }
    \label{fig:shapes}
\end{figure} 

\textbf{Merged Shapes:} 
We combine two regular polygons, such as a triangle and a square, where the shapes share a side. A simple shared-side algorithm can be applied to obtain the number of the sides: $s_{1} + s_{2} - 2$ where $s_{1}$ and $s_{2}$ are the number of sides of shape 1 and shape 2, respectively. Applying this algorithm to a triangle and a square gives: \(3 + 4 - 2 = 5\) total sides. Table~\ref{tab:composite_shapes} shows that models generally fail to implement this algorithm, with most achieving 0\% accuracy on merged shapes (Tri-Sq through Sq-Hep). LLaVA-OneVision and  LLaMA-3.2 each achieve over 90\% accuracy on exactly one merged shape. This is because most open-source models (except for Molmo, which is trained on counting) tend to guess a single value from 3 to 8 on every shape, thus getting ``guessing correctly'' on a specific category while scoring 0\% on all others. These results demonstrate that models do not implement the $s_{1} + s_{2} - 2$ algorithm.
    \begin{figure*}[htbp]
        \centering
        \includegraphics[width=\textwidth]{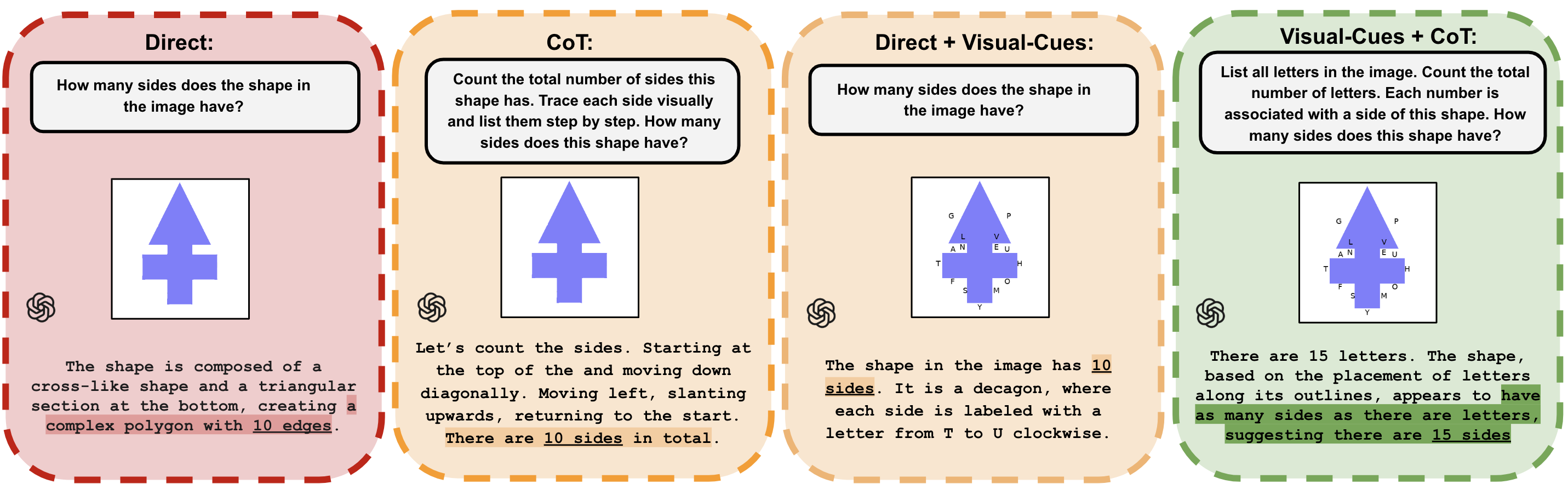} 
        \caption{Example outputs from GPT-4-Turbo on ``random letters'' annotations. }
        \label{fig:case}
    \end{figure*}
    
\textbf{Irregular Polygons:} Extending the experiment from Section~\ref{exp1_mllms}, we evaluate side-counting on irregular polygons with 3 to 8 sides. Performance remains low across all models, with GPT-4o, Molmo, and Math-LLaVA achieving the highest accuracy (42\%). Most correct predictions come from irregular triangles, while shapes with 4–8 sides yield near 0\% accuracy, suggesting poor generalization from regular to irregular polygons. 
   
\textbf{Abstract Shapes:} Arrows, stars, and plus signs are common in documents and presentations, thus more likely to appear in pretraining data. GPT-4o and GPT-4 Turbo achieve 84–92\% accuracy on these shapes, while open-source models score close to 0\%. When these shapes are combined into novel configurations (an arrow on top of a plus sign seen in Figure \ref{fig:example_first_page}), all models yield 0\% accuracy.

The results in Table \ref{tab:composite_shapes} highlight the influence of pretraining exposure, as models fail to generalize side counting beyond familiar shapes. Instead, their performance confirms our hypothesis that MLLMs rely on System 1 thinking, memorizing shape-side relationships rather than engaging System 2 reasoning to count and interpret visual information. The reliance on memorization over reasoning suggests a need for strategies that can shift MLLMs toward more deliberate, step-by-step problem-solving. 
 
\subsection{Visually-Cued CoT}
\begin{table}[h]
    \centering
    \scriptsize
    \resizebox{\columnwidth}{!}{
    \begin{tabular}{lcccccccc}
        \toprule
        Image Labels & \textbf{Qwen2-VL} & \textbf{Molmo} & \textbf{Janus-Pro} & \textbf{GPT-4-Turbo} & \textbf{GPT-4o} \\
        \midrule
        & & \textbf{Heptagon: Direct / VC-CoT} & & & \\            
        \midrule
        
        Numbers 1-7 & 93\% / \colorbox{lightblue}{\textbf{100\%}} & 60\% / \colorbox{lightblue}{69\%} & 1\% / \colorbox{lightblue}{\textbf{100\%}} & 79\% / \colorbox{lightblue}{98\%} & 43\% / \colorbox{lightblue}{\textbf{100\%}} \\
        Random Nums & 6\% / \colorbox{lightblue}{98\%} & 17\% / \colorbox{lightblue}{29\%} & 3\% / \colorbox{lightblue}{90\%} & 1\% / \colorbox{lightblue}{94\%} & 31\% / \colorbox{lightblue}{\textbf{100\%}} \\
        Letters A-G & 32\% / \colorbox{lightblue}{92\%} & 63\% / \colorbox{lightblue}{75\%} & 15\% / \colorbox{lightblue}{97\%} & 99\% / \colorbox{lightblue}{\textbf{100\%}} & 59\% / \colorbox{lightblue}{\textbf{100\%}} \\
        Random Letters & 0\% / \colorbox{lightblue}{83\%} & 49\% / \colorbox{lightblue}{\textbf{89\%}} & 7\% / \colorbox{lightblue}{74\%} & 7\% / \colorbox{lightblue}{82\%} & 7\% / \colorbox{lightblue}{93\%} \\
        \midrule
        No Annotations & 0\% / 99\%$^{\dagger}$  & 1\% / 1\%$^{\dagger}$ & 4\% / 14\%$^{\dagger}$  & 1\%/ 3\%$^{\dagger}$  & 22\% / 41\%$^{\dagger}$  \\
        \midrule
        & & \textbf{Arrow on Plus: Direct / VC-CoT} & & \\
        \midrule
        Numbers 1-15 & 29\% / \colorbox{lightblue}{\textbf{82\%}} & 0\% / \colorbox{lightblue}{\textbf{61\%}} & 13\% / \colorbox{lightblue}{\textbf{92\%}} & 97\% / \colorbox{lightblue}{\textbf{99\%}} & 88\% / \colorbox{lightblue}{\textbf{99\%}} \\
        Random Nums & 0\% / \colorbox{lightblue}{5\%} & 0\% / \colorbox{lightblue}{5\%} & 0\% / \colorbox{lightblue}{0\%} & 1\% / \colorbox{lightblue}{78\%} & 3\% / \colorbox{lightblue}{77\%} \\
        Letters A-O & 0\% / \colorbox{lightblue}{3\%} & 0\% / \colorbox{lightblue}{1\%} & 0\% / \colorbox{lightblue}{2\%} & 75\% / \colorbox{lightblue}{74\%} & 14\% / \colorbox{lightblue}{\textbf{99\%}} \\
        Random Letters & 0\% / \colorbox{lightblue}{0\%} & 0\% / \colorbox{lightblue}{2\%} & 0\% / \colorbox{lightblue}{1\%} & 0\% / \colorbox{lightblue}{25\%} & 3\% / \colorbox{lightblue}{96\%} \\
        \midrule
        No Annotations & 0\% / 0\%$^{\dagger}$  & 0\% / 0\%$^{\dagger}$  & 0\% / 0\%$^{\dagger}$  & 0\% / 1\%$^{\dagger}$  & 0\% / 1\%$^{\dagger}$  \\
        \bottomrule
    \end{tabular}
}
    \caption{Performance on heptagons and arrow on plus for Qwen2-VL, Molmo, Janus-Pro, and GPT models. Results are presented as direct / \colorbox{lightblue}{VC-CoT}. When there are no image annotations, we use a standard CoT prompt denoted by $\dagger$. VC-CoT is more effective than direct CoT prompting for MLLMs in side counting.}
    \label{tab:visual_cot}
\end{table}

Our results 
demonstrate that MLLMs heavily rely on System 1 thinking when completing visual-geometry tasks. We explore a potential solution to encourage System 2 reasoning in MLLMs by introducing Visually-Cued Chain-of-Thought (VC-CoT) prompting. CoT prompting is widely recognized for improving logical reasoning in LLMs by guiding models to break down complex tasks into intermediate steps \cite{cot}. However, prior work suggests that MLLMs struggle with CoT reasoning \cite{vlm_cot, mavis, mathvision}. Existing CoT frameworks for MLLMs focus primarily on the language model without leveraging information available in the image. VC-CoT incorporates explicit ``visual cues'' into CoT prompts to help bridge the gap between CoT success in LLMs and its effectiveness in MLLMs. 

We conduct a case study on two shapes from our previous experiments: a regular heptagon and an arrow on top of a plus sign. For both shapes, models faced significant difficulty in accurately counting their sides, even when the shapes were correctly identified as a ``heptagon'' or a ``triangular shape on a cross''. We consider three conditions: (1) plain images with no annotations, (2) images annotated with ordered numbers or random numbers, and (3) images annotated with ordered letters or random letters. For each case, we compared direct prompts (e.g., ``How many sides does the shape in the image have?'') with CoT prompts that explicitly reference visual annotations (see Figure \ref{fig:case}). We use random and ordered letters/numbers when annotating images to ensure that performance is not solely measured under conditions that provide explicit references to the number of sides. 
Even for plain images with no labels, we included a CoT prompt (seen in Figure \ref{fig:case}) to test whether reasoning prompts alone could enhance performance.

Table \ref{tab:visual_cot} shows the performance of Qwen2-VL, Molmo, Janus-Pro, GPT-4-Turbo, and GPT-4o, and Appendix Section \ref{app:vc_cot_full_tables} shows results for all other models. Molmo and Janus-Pro were explicitly designed for CoT reasoning through dedicated training, while GPT models are well-documented to excel with CoT prompting \cite{cot}, making them the most relevant candidates for VC-CoT. In Table \ref{tab:visual_cot}, we see that these models exhibit substantial improvements when visually cued CoT prompts are paired with explicit visual annotations, such as numbering or labeling each side. This is especially evident in the GPT models, where GPT-4o increases from 7\% to 93\%, in the random letters annotation. 

Across all models, note that annotations alone or CoT without annotations fail to override visual biases. A striking failure emerges even when models are explicitly given the correct answer through labeled sides (e.g., numbering a heptagon’s sides from 1-7). Non-GPT models still fail without CoT, exposing a fundamental weakness in visual perception. Without VC-CoT prompts, MLLMs fail to extract fine-grained details and default to System 1 memorized associations, hindering structured reasoning. However, linking reasoning steps to visual cues helps mitigate visual biases in these models.

\begin{table}[h]
    \centering
    \scriptsize
    \resizebox{\columnwidth}{!}{
    \begin{tabular}{lccc}
        \toprule
        Model & Direct & MathVerse CoT & VC-CoT \\
        \midrule
        Qwen2-VL  &  35.76\% &  41.11\% &  \textbf{41.95}\% \\
        Molmo & 33.33\% & 33.33\% & \textbf{38.32\%} \\
        Janus-Pro & 30.60\% & 29.66\% & \textbf{33.96\%} \\
        GPT-4-Turbo & 45.76\% & 48.01\% & \textbf{52.41\%} \\
        GPT-4o & 52.12\% & 55.35\% & \textbf{59.94\%} \\
        \midrule
        Average & 39.51\% & 41.49\% &  \textbf{45.32\%} \\ 
        
        \bottomrule
    \end{tabular}
    }
    \caption{\textbf{VC-CoT improves performance on MathVerse}. Compared to standard CoT prompting, explicitly guiding models to extract visual information leads to consistent performance improvements.}
    \label{tab:visual_cot_mathverse}
\end{table}

\subsection{Does VC-CoT generalize beyond side-counting?}

Given the success of VC-CoT prompts in improving side counting, we evaluate VC-CoT on a complex mathematical reasoning task. Many multimodal geometry datasets already include diagrams annotated with letters, numbers, and angles, providing a natural opportunity to leverage these visual cues. In particular, the vision-dominant split of the MathVerse dataset contains sufficient visual annotations to answer the question. While MathVerse includes CoT prompting, it only slightly modifies the question, changing ``please directly answer'' to ``please first conduct reasoning'', without guiding models to extract the available visual information. In contrast, our visually-cued CoT explicitly prompts models to first identify all present shapes, numbers, and letters and establish their spatial and numerical relationships before answering the question. This structured approach aims to engage System 2 reasoning, forcing models to process the visual structure, retrieve relevant information from the image, and then reason about their relationship. 

As shown in Table~\ref{tab:visual_cot_mathverse}, specifically addressing the visual annotations through VC-CoT improves accuracy across all models. For Molmo and Janus-Pro, MathVerse CoT, does not increase performance, and even decreases compared to the direct prompt. As seen in Appendix Section \ref{app:vc_cot_mathverse}, these results generalize to other models as well. While these experiments serve as a case study, they highlight a promising direction for enhancing the reasoning capabilities of MLLMs. We believe incorporating visual cues in CoT prompts can enhance MLLMs' ability to bridge the gap between vision and language reasoning.

\section{Conclusion}
Our study highlights fundamental limitations in MLLMs' ability to integrate visual information for reasoning. While LLM backbones possess strong geometric knowledge, vision encoders remain the primary bottleneck, forcing MLLMs to rely on System 1 thinking instead of System 2 reasoning. Even in simple tasks like shape identification and side counting, MLLMs default to memorized patterns rather than systematically analyzing visual inputs. As an initial step to improving MLLM reasoning, we introduce Visually-Cued Chain-of-Thought (VC-CoT) prompting to engage System 2 reasoning. Our results show that explicitly guiding models to extract and reason about visual cues improves performance, boosting GPT-4o's accuracy by 86\% on our shape dataset as well as enhancing results on MathVerse.

Our findings serve as a broader call to action for the MLLM research community. To effectively pinpoint and address limitations in vision-language integration, we emphasize the need to explore simple, controlled scenarios before engaging with complex multimodal benchmarks. By improving fine-grained visual perception and utilizing visual cues for prompting, researchers can move toward MLLMs that engage in true vision-based System 2 reasoning rather than defaulting to LLM-driven deduction.

\section{Limitations}
Although we evaluated a diverse range of 13 models, including open-sourced, closed-sourced, and fine-tuned models specialized in mathematical reasoning, there are two recent models we were unable to assess. At the time of writing, the weights for MAVIS \cite{mavis} are not publicly available. Additionally, SVE-Math \cite{sve-math}, which adapts the vision encoder with a GeoGLIP encoder trained on mathematical diagrams, shows promise for mathematical reasoning. However, there are unresolved issues in the code base, including missing files in the official GitHub repository, which have yet to be addressed. We will add these models to our evaluation suite as soon as these issues are addressed. 

Furthermore, our study is designed around a controlled synthetic environment, allowing us to systematically analyze the components of visual geometry datasets. Thus, our method, Visually-Cued Chain-of-Thought (VC-CoT), is particularly geared for datasets that include image annotations, such as Mathverse \cite{zhang2024mathversedoesmultimodalllm}. Future work could explore how VC-CoT adapts to real-world images, potentially leveraging other types of annotations, such as pointing data from Pixmo \cite{molmo} for example, to bridge the gap between synthetic and real-world settings.

\bibliography{bib}

\newpage
\appendix

\section{Model Selection and Details}\label{app:models}

We evaluate a diverse set of multimodal language models (MLLMs) spanning different architectures, vision encoders, and parameter scales. Table~\ref{tab:model_details} provides details on the open-source models used in our experiments. These models are sourced from Hugging Face, with their specific repository paths listed for reproducibility.

In addition to open-source models, we also evaluate proprietary models using the OpenAI API. Specifically, we test \texttt{gpt-4o} and \texttt{gpt-4-turbo}, for comparison against open-source alternatives. Since their exact architectures and vision encoders are not publicly disclosed, we cannot report model details.

\begin{table*}[h]
    \centering
    \small
    \resizebox{\textwidth}{!}{
    \begin{tabular}{lcccc}
        \toprule
        \textbf{Model Name} & \textbf{Underlying LM} & \textbf{Vision Encoder} & \textbf{Size} & \textbf{HuggingFace Path} \\
        \midrule
        LLaVA-1.5 & Vicuna & CLIP (ViT-L/14) & 7B & \texttt{llava-hf/llava-1.5-7b-hf} \\
        LLaVA-Next (LLaVA-1.6) & Mistral & CLIP (ViT-L/14) & 7B & \texttt{llava-hf/llava-v1.6-mistral-7b-hf} \\
        LLaVA-OneVision & Qwen-2 & SigLIP & 7B & \texttt{llava-hf/llava-onevision-qwen2-7b-ov-hf} \\
        Qwen2-VL & Qwen2 & DFN-ViT w/ RoPE-2D & 7B & \texttt{Qwen/Qwen2-VL-7B-Instruct} \\
        InternVL & QLLaMA & InternViT & 8B & \texttt{OpenGVLab/InternVL2-8B} \\
        Molmo & OLMo & MetaCLIP & 7B & \texttt{cyan2k/molmo-7B-D-bnb-4bit} \\
        Janus Pro & DeepSeek-LLM & SigLIP-Large-Patch16-384 & 7B & \texttt{deepseek-ai/Janus-Pro-7B} \\
        \midrule
        Math-LLaVA & Vicuna & CLIP (ViT-L/14) & 13B & \texttt{Zhiqiang007/Math-LLaVA} \\
        G-LLaVA & Vicuna & CLIP (ViT-L/14) & 7B & \texttt{renjiepi/G-LLaVA-7B} \\
        Math-PUMA & Qwen2 & DFN-ViT w/ RoPE-2D & 7B & \texttt{Math-PUMA/Math-PUMA\_Qwen2VL-7B} \\
        \bottomrule
    \end{tabular}
    }
    \caption{Details of evaluated models, including underlying language models, vision encoders, sizes, and Hugging Face model paths.}
    \label{tab:model_details}
\end{table*}

\section{Image Generation}\label{sec:image_gen}
\subsection{One-Shape Images}\label{app:one_shape}
\begin{figure}[h!]
    \centering
    \includegraphics[width=0.7\columnwidth]{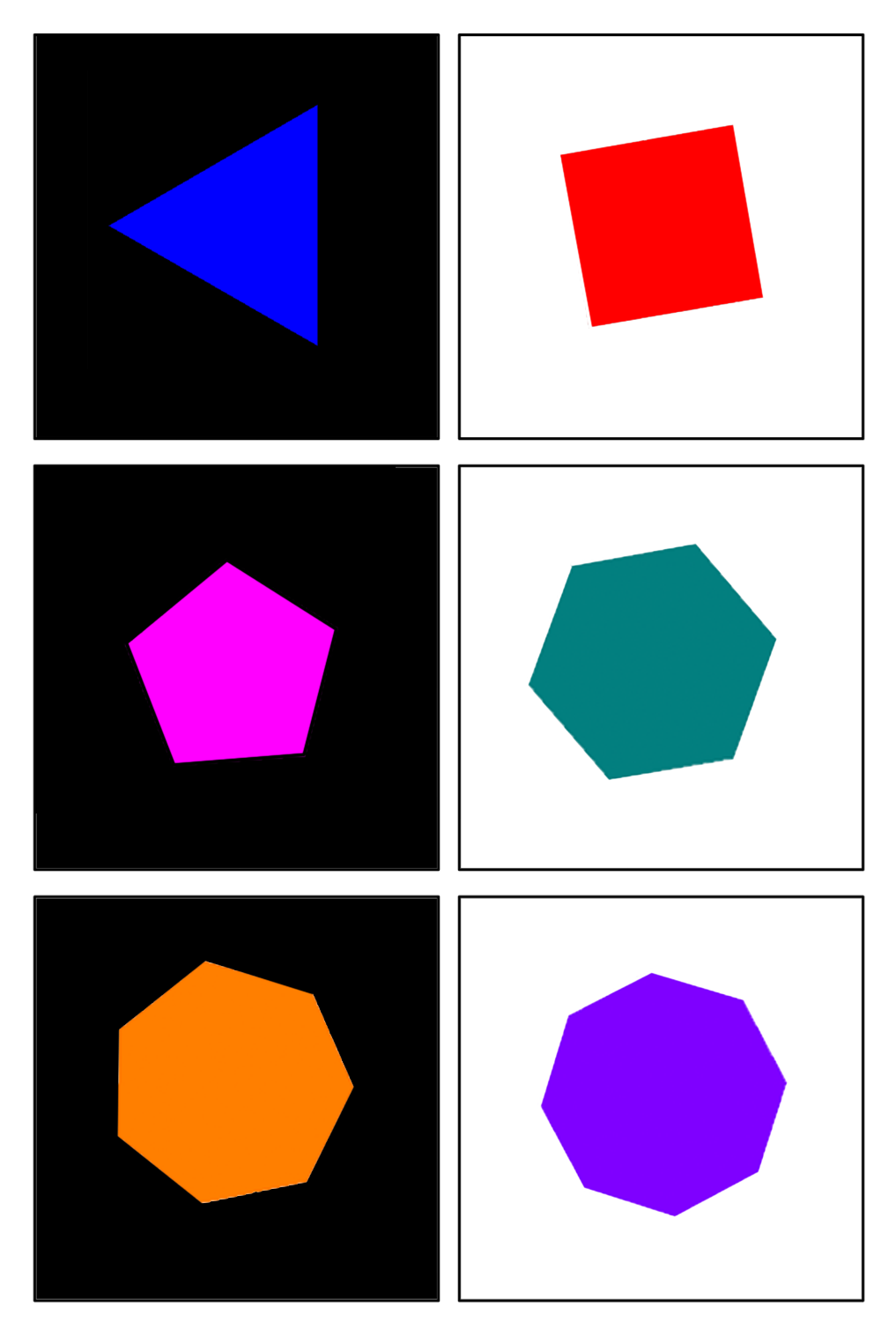}\caption{Examples of regular polygons (triangle through octagon) used in our evaluation in Section \ref{exp1_mllms}.}
    \label{fig:regular_polygons}
\end{figure}

To create a diverse dataset of geometric shapes, we generate images of squares, equilateral triangles, pentagons, hexagons, heptagons, octagons, and circles. Each shape is rendered in various sizes, orientations, and colors. Specifically, we vary the background color (white, black, red, blue) and fill color, selecting from a palette of 20 distinct hues. We ensured that when the background color equaled the shape color, those samples were not used. Additionally, each shape is rotated in steps of 10 degrees, covering a full 360-degree range. We account for shape symmetry, so we do not have duplicate samples on equivalent rotation angles (e.g., 90 degrees and 180 degrees when rotating a square). Finally, we vary in 9 different shape sizes. For the results in Table \ref{tab:combined_shape_sides_accuracy}, we randomly sample a subset of \textbf{2,000} images to ensure diversity while maintaining computational efficiency. Figure \ref{fig:regular_polygons} shows examples from our dataset.

\begin{figure}
    \centering
    \includegraphics[width=\columnwidth]{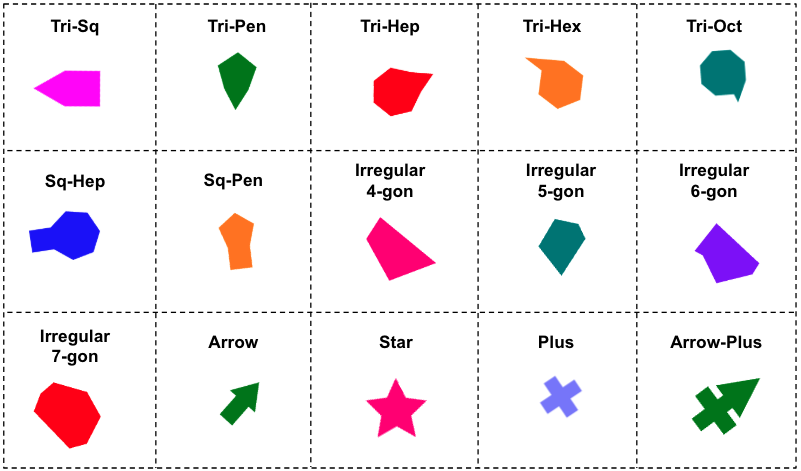}
    \caption{Examples of all merged, irregular and abstract shapes used to create Table~\ref{tab:composite_shapes}}
    \label{fig:enter-label}
\end{figure}
\subsection{Two-Shape Images}\label{app:two_shape}
In addition to single-shape images, we generate images containing two distinct geometric shapes. Each image consists of a randomly selected pair from squares, equilateral triangles, pentagons, hexagons, heptagons, and octagons, placed in non-overlapping positions on a $400 \times 400$ canvas. Shapes vary in size, rotation, and color, with sizes randomly chosen from $\{80, 90\}$ pixels and rotations sampled in $10^\circ$ increments. Each shape is assigned a fill color from a set of 20 hues, while the background color is randomly selected from white, black, red, or blue. In total, we generated \textbf{5,090} unique images, and used all of them for evaluation.

\subsection{Abstract Shape Image Generation}\label{app:abstract_shapes}

\textbf{Merged shapes.} For the results in Table \ref{tab:composite_shapes}, we generate images containing two connected geometric shapes. We consider ten unique shape combinations, including Triangle-Square, Pentagon-Square, Hexagon-Triangle, and Octagon-Square, ensuring a range of structural complexities. Each combination is assigned a total side count based on the sum of its components. Shapes are placed in a connected configuration, rotated at random angles, and rendered with diverse colors on varying background shades. In total, \textbf{100} images for the merged shape category.

\textbf{Irregular Polygons.} We generate images of abstract irregular polygons with between 5 and 8 sides. Each polygon is randomly positioned on a $400 \times 400$ canvas with a randomly assigned background color (white, black, red, or blue) and a distinct shape color chosen from 20 hues. The polygon vertices are generated with constraints to ensure reasonable side lengths and angles, avoiding excessively sharp or flat corners. The number of sides is correctly recorded in the filenames to ensure metadata accuracy. In total, we generate \textbf{200} unique irregular polygons for analysis.

\textbf{Common Shapes.}
We consider four shape configurations: plus sign (or a cross), star, arrow, and arrow on plus sign (or a triangle on top of a cross). Each shape undergoes a uniform scaling transformation and is randomly rotated between $0^\circ$ and $360^\circ$. This process results in a dataset of \textbf{200} uniquely transformed composite shapes.

\section{What does the text-decoder know?}
\begin{table}[h]
    \centering
    \scriptsize
    \resizebox{\columnwidth}{!}{
    \begin{tabular}{lc}
        \toprule
        \multicolumn{2}{c}\textbf{Name the \textit{ \textless n \textgreater}-sided polygon / Num of sides in \textless \textit{shape} \textgreater} \\
        \midrule
        LM Only & Accuracy (\%) \\
        \midrule
        LLaVA-1.5 & 67\% / 100\% \\
        LLaVA-Next & 100\% / 100\% \\
        LLaVA-OneVision & 100\% / 100\% \\
        Qwen2-VL & 100\% / 100\% \\
        LLaMA-3.2 & 100\% / 100\% \\
        InternVL & 100\% / 100\% \\
        Molmo & 100\% / 100\% \\
        Janus-Pro & 100\% / 100\% \\
        \midrule
        Math-LLaVA & 100\% / 100\% \\
        G-LLaVA & 50\% / 100\% \\
        Math-PUMA & 100\% / 100\% \\
        \midrule
        GPT-4o & 100\% / 100\% \\
        GPT-4-Turbo & 100\% / 100\% \\
        \bottomrule
    \end{tabular}
    }
    \caption{Accuracy of the LLM-backbone on the polygon naming and side counting tasks: 
    ``What is the name of a \textit{\textless n\textgreater}-sided regular polygon? / 
    How many sides does \textit{\textless shape\textgreater} have?'' for $n \in \{3,4,...,8 \}$ and shape in $\{$triangle, square,..., octagon$\}$.} 
    \label{tab:LM_only}
\end{table}

To evaluate LLMs' recall of basic geometric properties, we pose two sets of six questions: (1) naming a polygon given its number of sides (3–8) and (2) identifying the number of sides in a named shape. As shown in Appendix Table~\ref{tab:LM_only}, models perform nearly perfectly, with all achieving 100\% accuracy on the second task and 11 out of 13 models doing so on the first. The exceptions are LLaVA-1.5, which misclassified a heptagon as a hexagon and a square as a tetrahedron, and G-LLaVA, which answered “regular polygon” instead of the specific shape name for several cases.

\section{What does the vision-encoder see?}\label{app:vision_encoder_experiments}
\begin{figure*}[h!]
    \centering
    \includegraphics[width=\textwidth]{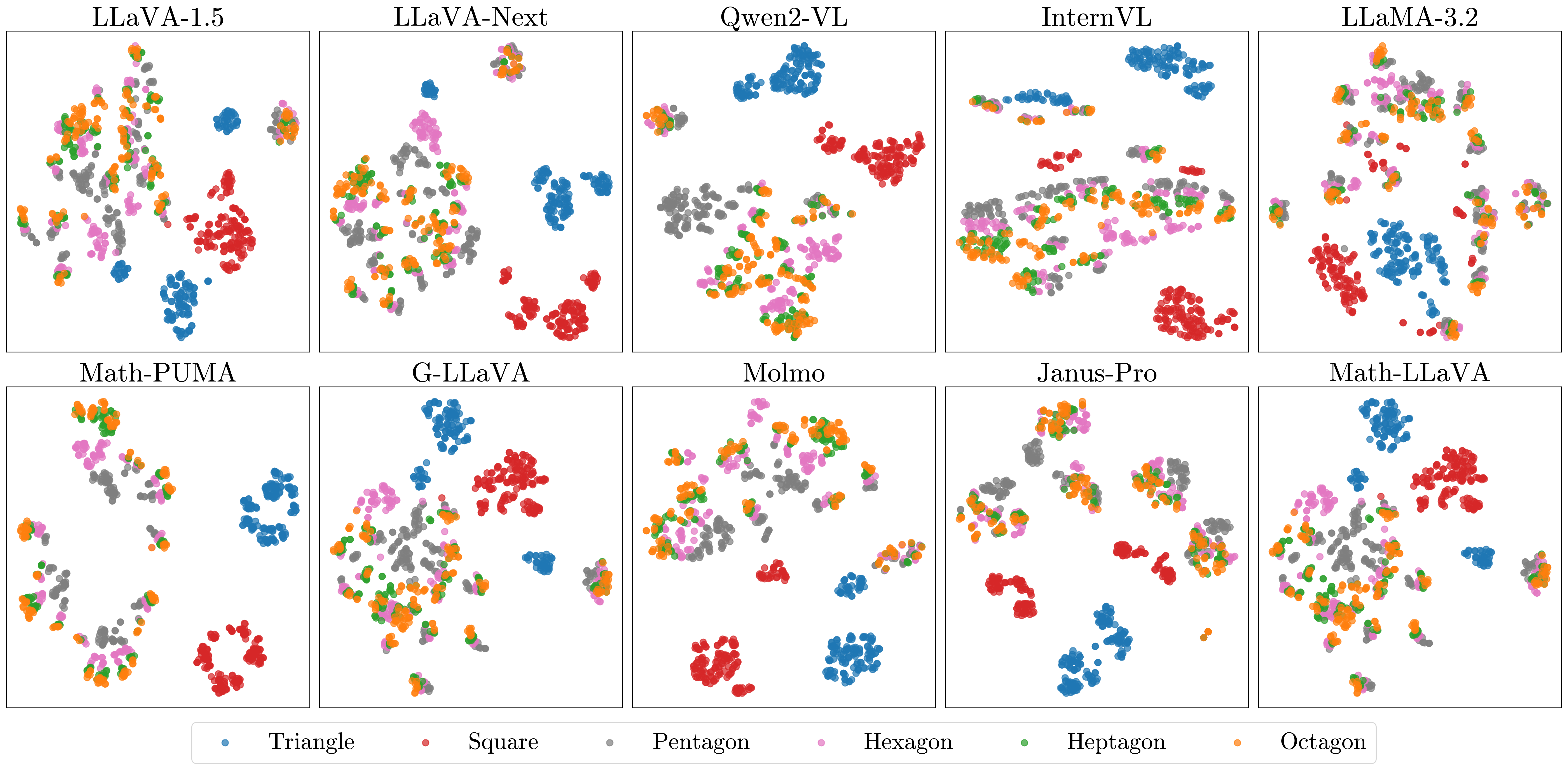} 
    \caption{t-SNE plots from the vision encoders of all models considered in this paper. For each image, we take the average over all image patches.}
    \label{fig:all_vision_tsne}
\end{figure*}

Figure~\ref{fig:all_vision_tsne} presents t-SNE visualizations of vision encoder embeddings across all models. Common shapes, such as triangles and squares, consistently form distinct clusters, whereas less frequent shapes, including hexagons, heptagons, and octagons, exhibit dispersed and overlapping embeddings. This confirms that vision encoders struggle to differentiate between less common shapes, leading to poor shape generalization.
Nearest neighbor analysis, in Table \ref{tab:nn_purity}, further supports this, showing that models reliably group common shapes but misclassify rarer ones, often embedding them into mixed clusters.

\begin{table*}[h]
    \centering
    \scriptsize
    \resizebox{0.8\textwidth}{!}{
    \begin{tabular}{lcccccc}
        \hline
        Model & Triangle & Square & Pentagon & Hexagon & Heptagon & Octagon \\
        \hline
        LLaVA-1.5 & 95\% & 95\% & 65\% & 50\% & 44\% & 48\% \\
        LLaVA-Next & 99\% & 97\% & 66\% & 58\% & 50\% & 46\% \\
        Qwen2-VL & 100\% & 98\% & 84\% & 67\% & 51\% & 50\% \\
        InternVL & 97\% & 90\% & 58\% & 50\% & 47\% & 47\% \\
        LLaVA-OneVision & 100\% & 100\% & 82\% & 58\% & 52\% & 52\% \\
        LLaMA-3.2 & 87\% & 79\% & 40\% & 37\% & 36\% & 34\% \\
        Math-PUMA & 100\% & 100\% & 76\% & 57\% & 51\% & 51\% \\
        G-LLaVA & 96\% & 97\% & 65\% & 48\% & 44\% & 47\% \\
        Molmo & 99\% & 97\% & 76\% & 54\% & 51\% & 46\% \\
        Janus-Pro & 98\% & 95\% & 74\% & 47\% & 46\% & 43\% \\
        Math-LLaVA & 96\% & 97\% & 65\% & 48\% & 44\% & 48\% \\
        \hline
    \end{tabular}
    }
    \caption{Average percentage of nearest neighbors of vision encoder embeddings that share the same shape label.}
    \label{tab:nn_purity}
\end{table*}
\begin{table}[h]
    \centering
    \scriptsize
    \resizebox{\columnwidth}{!}{
    \begin{tabular}{lrrr}
        \toprule
        {} &  Direct &  MathVerse CoT &  VC-CoT \\
        \midrule
        LLaVA-1.5  &  29.39\% &  29.97\% &  28.34\% \\
        LLaVA-Next  &  28.48\% &  23.46\% &  30.95\% \\
        LLaVA-OneVision  &  47.88\% &  48.62\% &  50.16\% \\
        Qwen2-VL       &  35.76\% &  41.11\% &  41.95\% \\
        InternVL   &  34.85\% &  47.09\% &  44.86\% \\
        LLaMA-3.2  &  22.73\% &  32.72\% &  35.20\% \\
        Math-LLaVA &  36.06\% &  34.86\% &  36.14\% \\
        G-LLaVA    &  35.76\% &  36.70\% &  35.83\% \\
        Math-PUMA  &  39.70\% &  36.09\% &  37.35\% \\
        \midrule
        Average    &  34.51\% &  36.73\% &  37.86\% \\
        \bottomrule
    \end{tabular}
    }
    \caption{Impact of Visually-Cued CoT (VC-CoT) on MathVerse across different models. Compared to standard CoT prompting, explicitly guiding models to extract and process visual information leads to consistent performance improvements. This table includes all models not in Table \ref{tab:visual_cot_mathverse}.}
    \label{tab:vc_cot_mathverse_app}
\end{table}
\section{Google N-Grams Analysis of Shape Frequencies}\label{app:ngrams_analysis}
\begin{figure}[htbp]
        \centering
        \includegraphics[width=0.9\columnwidth]{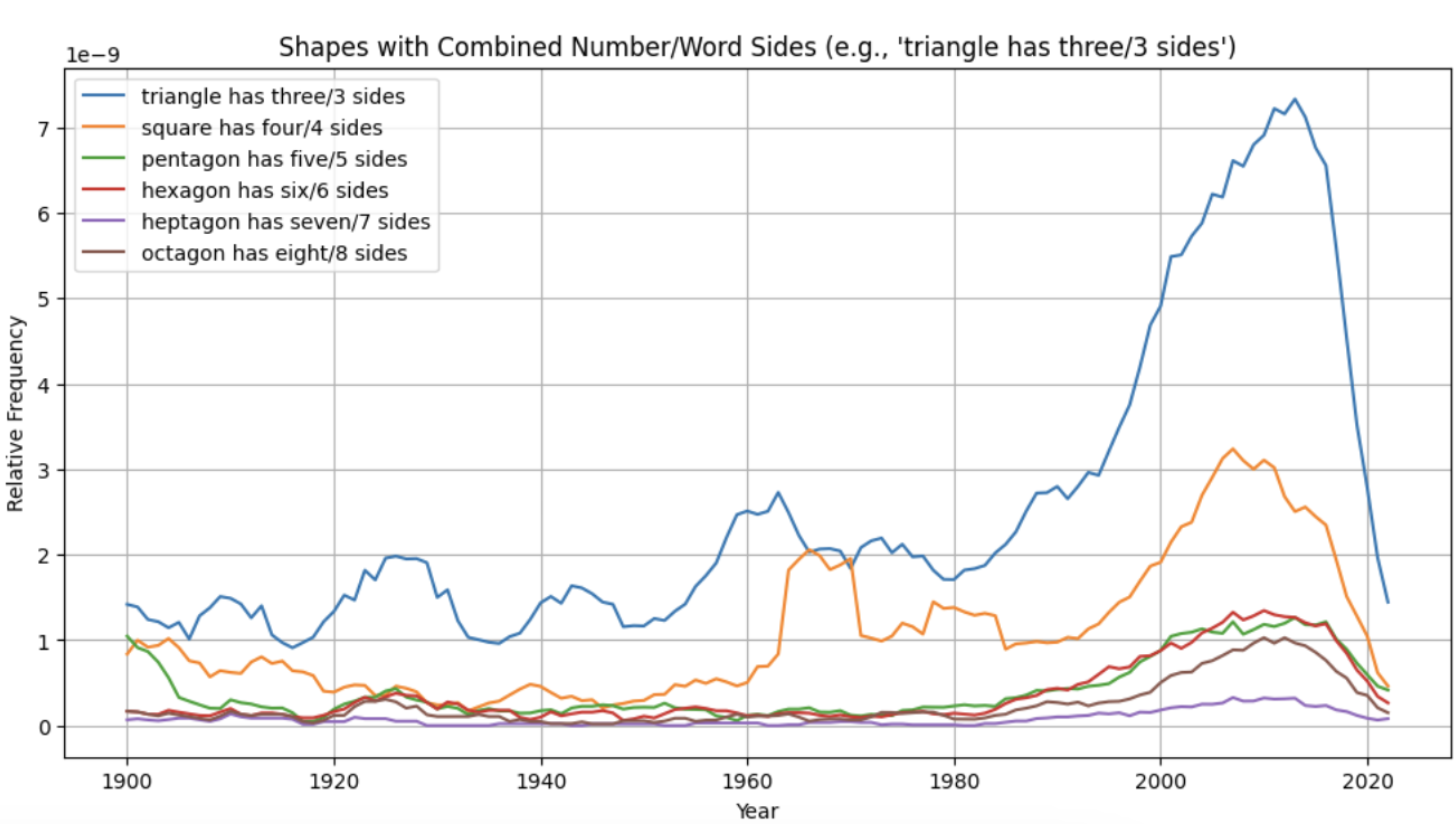} 
        \caption{Google Books N-Grams showing frequency of different shape mentions with their corresponding numbers over the years. There is a clear bias towards triangles and squares compared to the rest. `Heptagon'' is the only shape GPT-4-Turbo misclassified and it is the least frequent in this plot. }
        \label{fig:ngrams}
    \end{figure}

Figure~\ref{fig:ngrams} presents an analysis of shape mention frequencies over time using data from Google Books N-Grams. The plot highlights the relative prominence of different shapes and their corresponding numbers in written text. Notably, triangles and squares are mentioned significantly more frequently than other shapes, reflecting a clear bias in natural language usage.

This bias in textual representation may contribute to discrepancies in model performance, as models trained on large language corpora are more likely to encounter these common shapes during training. For instance, ``heptagon,'' the only shape misclassified by GPT-4-Turbo in our experiments, is also the least frequent shape in this plot. This suggests a possible correlation between linguistic frequency and model performance on geometric reasoning tasks.

\section{Molmo Demo Examples}\label{app:molmo_experiment}

To evaluate the pointing ability of Molmo, we experiment using the HuggingFace-hosted model as well as its interactive demo website. Specifically, we prompt the model three different times with variations of prompts closely matching examples in the Molmo paper. The task involves pointing to and identifying the sides of a simple black heptagon.

In all prompt variations, Molmo consistently points to the heptagon but incorrectly assigns the points, resulting in 10 sides instead of the correct 7. However, when the pointing task is omitted, and the prompt is modified to include the instruction \textit{``Answer with the number''}, Molmo correctly identifies the number of sides as 7 (which correspond to the results in Table \ref{tab:combined_shape_sides_accuracy}). This suggests that while Molmo performs reasonably well in direct counting tasks, its pointing mechanism struggles to accurately map geometric features in images.

\begin{figure*}[h!]
    \centering
    \includegraphics[width=\textwidth]{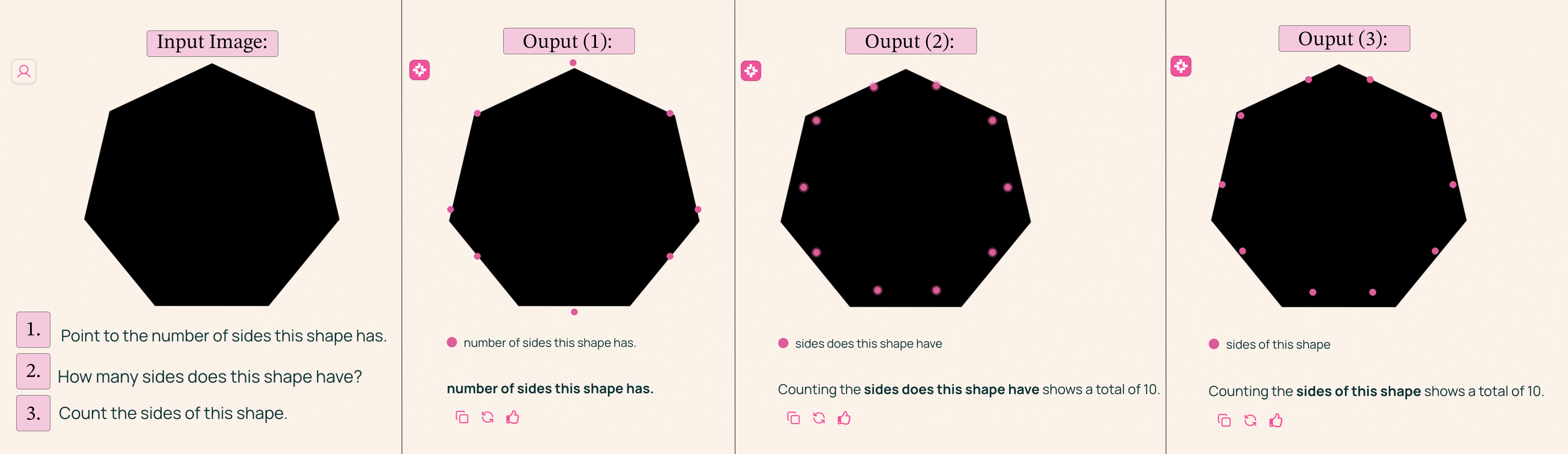} 
    \caption{Screenshots from Molmo's demo website. In all prompts, Molmo misassigns points, counting 10 sides instead of 7.}
    \label{fig:molmo_demo}
\end{figure*}

Figure~\ref{fig:molmo_demo} shows real screenshots taken from the Molmo demo website, illustrating its outputs for the shape-recognition and side-counting task.

\section{Visually-Cued Chain-of-Thought Shape Results}\label{app:vc_cot_full_tables}

Tables \ref{tab:vc-cot_hept_full_app} and \ref{tab:vc_cot_triangle_cross_app} present results for the remaining models not included in Table \ref{tab:visual_cot}.

Prior work has shown that standard Chain-of-Thought (CoT) prompting often fails to improve, and in some cases, even decreases mathematical reasoning performance in MLLMs \cite{mathvision}. This has been attributed to the lack of fine-grained visual information captured by most vision encoders, which struggle to integrate step-by-step reasoning with visual context \cite{vlm_cot, mavis}. Specifically, models trained on vision-language instruction datasets tend to prioritize short-answer responses, limiting their ability to perform multi-step reasoning \cite{vlm_cot}.

Our findings reinforce some of these observations. As seen in Table \ref{tab:vc-cot_hept_full_app}, most models struggle with side-counting on a heptagon, even with explicit ordered labels (1-7). VC-CoT, helps promote some correct predictions, but not in all models. We hypothesize that their rigid clustering patterns (Figure \ref{fig:cluster}) prevent them from adapting their reasoning when prompted. For example, LLaVA-OneVision misclassified a heptagon as a hexagon, stating: “The shape is a hexagon. The letters are A, B, C, D, E, F, and G. There are 7 letters in total... A hexagon has 6 sides.” These models fail to adjust their predictions even when provided with explicit step-by-step reasoning.

As shown in Table \ref{tab:vc_cot_triangle_cross_app}, some models exhibit different behaviors when reasoning about more complex composite shapes, such as the ``arrow on plus sign'' configuration. Notably, Math-PUMA's strong visual priors make VC-CoT ineffective for counting heptagon sides, where deeply ingrained biases interfere with reasoning. However, on far more complex shapes, like a 15-sided "arrow on plus sign," VC-CoT significantly improves performance. This is likely because the lack of a strong prior allows Math-PUMA to engage in step-by-step reasoning rather than relying on memorized heuristics. 
\section{VC-CoT Results on MathVerse}\label{app:vc_cot_mathverse}

Table \ref{tab:vc_cot_mathverse_app} presents the impact of Visually-Cued Chain-of-Thought (VC-CoT) prompting on MathVerse accuracy across models not included in Table \ref{tab:visual_cot_mathverse}. Prior work has demonstrated that standard Chain-of-Thought (CoT) prompting can be ineffective, and in some cases, even detrimental to mathematical reasoning performance in multimodal large language models (MLLMs) \cite{mathvision}. This is largely due to the training biases of vision-language models, which tend to favor short-answer responses over structured, multi-step reasoning \cite{vlm_cot}.

Despite these challenges, our results show that VC-CoT improves overall performance across these models, with an average accuracy increase from 34.51\% (Direct) to 37.86\% (VC-CoT). Notably, 9 out of 13 models (including ones in Table \ref{tab:visual_cot_mathverse}) benefit from visually guided reasoning, suggesting that explicit extraction and processing of visual information can help compensate for the weaknesses of generic CoT prompting. 

These findings suggest that while VC-CoT is not a universal solution, it provides a tangible benefit even for models not explicitly designed for reasoning through CoT. Future work may explore fine-tuning strategies that further leverage visual embeddings to enhance mathematical problem-solving capabilities in MLLMs.
\newpage
\begin{table*}[h]
    \centering
    \scriptsize
    \resizebox{\textwidth}{!}{
    \begin{tabular}{lcccccccccccccccccc}
        \toprule
        \multirow{2}{*}{Image Labels} & \multicolumn{2}{c}{LLaVA-1.5} & \multicolumn{2}{c}{LLaVA-Next} & \multicolumn{2}{c}{LLaVA-OneVision} & \multicolumn{2}{c}{Qwen2-VL} & \multicolumn{2}{c}{InternVL} & \multicolumn{2}{c}{LLaMA-3.2} & \multicolumn{2}{c}{Math-LLaVA} & \multicolumn{2}{c}{G-LLaVA} & \multicolumn{2}{c}{Math-PUMA} \\
        \cmidrule(lr){2-3} \cmidrule(lr){4-5} \cmidrule(lr){6-7} \cmidrule(lr){8-9} \cmidrule(lr){10-11} \cmidrule(lr){12-13} \cmidrule(lr){14-15} \cmidrule(lr){16-17} \cmidrule(lr){18-19}
         & Direct & VC-CoT & Direct & VC-CoT & Direct & VC-CoT & Direct & VC-CoT & Direct & VC-CoT & Direct & VC-CoT & Direct & VC-CoT & Direct & VC-CoT & Direct & VC-CoT \\
        \midrule
        Numbers 1-7 & 0\% & 1\% & 0\% & 5\% & 0\% & 17\% & 93\% & 100\% & 26\% & 68\% & 100\% & 100\% & 0\% & 0\% & 79\% & 81\% & 0\% & 0\% \\
        Random Numbers & 0\% & 0\% & 0\% & 0\% & 0\% & 0\% & 6\% & 98\% & 1\% & 21\% & 32\% & 50\% & 0\% & 0\% & 2\% & 0\% & 0\% & 1\% \\
        Letters A-G & 0\% & 0\% & 0\% & 0\% & 0\% & 0\% & 32\% & 92\% & 1\% & 2\% & 45\% & 96\% & 0\% & 0\% & 0\% & 0\% & 0\% & 0\% \\
        Random Letters & 0\% & 0\% & 0\% & 0\% & 0\% & 0\% & 0\% & 83\% & 0\% & 3\% & 3\% & 58\% & 0\% & 0\% & 0\% & 0\% & 0\% & 1\% \\
        \midrule
         & Direct & CoT & Direct & CoT & Direct & CoT & Direct & CoT & Direct & CoT & Direct & CoT & Direct & CoT & Direct & CoT & Direct & CoT\\
        \cmidrule(lr){2-3} \cmidrule(lr){4-5} \cmidrule(lr){6-7} \cmidrule(lr){8-9} \cmidrule(lr){10-11}
        \cmidrule(lr){12-13}
        \cmidrule(lr){14-15}
        \cmidrule(lr){16-17}
        \cmidrule(lr){18-19}
        No Annotations & 0\% & 0\% & 0\% & 0\% & 0\% & 0\% & 12\% & 99\% & 0\% & 0\% & 0\% & 0\% & 0\% & 0\% & 0\% & 0\% & 0\% & 0\% \\
        \bottomrule
    \end{tabular}
    }
    \caption{Performance of Visually-Cued Chain-of-Thought (VC-CoT) prompting on heptagons. This table includes models not used in Table \ref{tab:visual_cot}.}
    \label{tab:vc-cot_hept_full_app}
\end{table*}

\begin{table*}[h]
    \centering
    \scriptsize
    \resizebox{\textwidth}{!}{
    \begin{tabular}{lcccccccccccccccccc}
        \toprule
        \multirow{2}{*}{Image Labels} & \multicolumn{2}{c}{LLaVA-1.5} & \multicolumn{2}{c}{LLaVA-Next} & \multicolumn{2}{c}{LLaVA-OneVision} & \multicolumn{2}{c}{Qwen2-VL} & \multicolumn{2}{c}{InternVL} & \multicolumn{2}{c}{LLaMA-3.2} & \multicolumn{2}{c}{Math-LLaVA} & \multicolumn{2}{c}{G-LLaVA} & \multicolumn{2}{c}{Math-PUMA} \\
        \cmidrule(lr){2-3} \cmidrule(lr){4-5} \cmidrule(lr){6-7} \cmidrule(lr){8-9} \cmidrule(lr){10-11} \cmidrule(lr){12-13} \cmidrule(lr){14-15} \cmidrule(lr){16-17} \cmidrule(lr){18-19}
         & direct & VC-CoT & direct & VC-CoT & direct & VC-CoT & direct & VC-CoT & direct & VC-CoT & direct & VC-CoT & direct & VC-CoT & direct & VC-CoT & direct & VC-CoT \\
        \midrule
        Numbers 1-15 & 0\% & 8\% & 12\% & 28\% & 26\% & 26\% & 29\% & 82\% & 0\% & 14\% & 99\% & 99\% & 0\% & 0\% & 0\% & 0\% & 0\% & 61\% \\
        Random Numbers & 0\% & 0\% & 0\% & 0\% & 0\% & 0\% & 0\% & 5\% & 0\% & 3\% & 32\% & 21\% & 0\% & 0\% & 0\% & 0\% & 0\% & 28\% \\
        Letters A-O & 0\% & 1\% & 0\% & 0\% & 0\% & 1\% & 0\% & 3\% & 0\% & 1\% & 0\% & 62\% & 0\% & 0\% & 0\% & 0\% & 0\% & 14\% \\
        Random Letters & 0\% & 0\% & 0\% & 0\% & 0\% & 0\% & 0\% & 0\% & 0\% & 1\% & 0\% & 84\% & 0\% & 0\% & 0\% & 0\% & 0\% & 0\% \\
        \midrule
         & Direct & CoT & Direct & CoT & Direct & CoT & Direct & CoT & Direct & CoT & Direct & CoT & Direct & CoT & Direct & CoT & Direct & CoT\\
        \cmidrule(lr){2-3} \cmidrule(lr){4-5} \cmidrule(lr){6-7} \cmidrule(lr){8-9} \cmidrule(lr){10-11}
        \cmidrule(lr){12-13}
        \cmidrule(lr){14-15}
        \cmidrule(lr){16-17}
        \cmidrule(lr){18-19}
        No Annotations & 0\% & 0\% & 0\% & 0\% & 0\% & 0\% & 0\% & 0\% & 0\% & 0\% & 0\% & 0\% & 0\% & 0\% & 0\% & 0\% & 0\% & 0\% \\
        \bottomrule
    \end{tabular}
    }
    \caption{Performance of Visually-Cued Chain-of-Thought (VC-CoT) prompting on ``arrow on top of plus sign'' shape. This table includes models not used in Table \ref{tab:visual_cot}.}
    \label{tab:vc_cot_triangle_cross_app}
\end{table*}

\end{document}